\documentclass[a4paper]{article}
\usepackage[latin1]{inputenc}
\usepackage[T1]{fontenc}
\usepackage{pgf}
\usepackage{url}
\usepackage[english]{babel}
\usepackage{multirow}
\usepackage{fancyhdr}
\usepackage{anysize}
\usepackage{amssymb}
\usepackage{graphicx}
\usepackage[latin1]{inputenc}
\usepackage{url}
\usepackage{algorithm}
\usepackage{algpseudocode}
\usepackage[sort, numbers]{natbib}
\usepackage{ntheorem}
\usepackage{amsmath}
\usepackage{amstext}

\begin{document}

\title{\bf{EigenEvent: An Algorithm for Event Detection from Complex Data streams in Syndromic Surveillance}}
\author{Hadi Fanaee-T and João Gama}
\date{}
\maketitle
\begin{center}
Laboratory of Artificial Intelligence and Decision Support (LIAAD), University of Porto\\
INESC TEC, Rua Dr. Roberto Frias, Porto, Portugal\\
hadi.fanaee@fe.up.pt and jgama@fep.up.pt\\
\end{center}

\begin{abstract}

Syndromic surveillance systems continuously monitor multiple pre-diagnostic daily streams of indicators from different regions with the aim of early detection of disease outbreaks. The main objective of these systems is to detect outbreaks hours or days before the clinical and laboratory confirmation. The type of data that is being generated via these systems is usually multivariate and seasonal with spatial and temporal dimensions. The algorithm What's Strange About Recent Events (WSARE) is the state-of-the-art method for such problems. It exhaustively searches for contrast sets in the multivariate data and signals an alarm when find statistically significant rules. This bottom-up approach presents a much lower detection delay comparing the existing top-down approaches. However, WSARE is very sensitive to the small-scale changes and subsequently comes with a relatively high rate of false alarms. We propose a new approach called EigenEvent that is neither fully top-down nor bottom-up. In this method, we instead of top-down or bottom-up search, track changes in data correlation structure via eigenspace techniques. This new methodology enables us to detect both overall changes (via eigenvalue) and dimension-level changes (via eigenvectors). Experimental results on hundred sets of benchmark data reveals that EigenEvent presents a better overall performance comparing state-of-the-art, in particular in terms of the false alarm rate.
\end{abstract}

{\bf Keywords:} Event Detection, Complex Data Streams, Tensor Decomposition, Syndromic Surveillance

\section{Introduction}\label{tab:introduction}

The goal of syndromic surveillance systems is to enable earlier detection of epidemics and a more timely public health response, hours or days before clinical and laboratory confirmation comes out \cite{henning2004overview}. Two kinds of events are usually required to be detected: man-made events such as bio-terrorist activities like anthrax attacks \cite{gursky2003anthrax} and natural events such as epidemic diseases like H1N1, avian influenza, SARS, and West Nile Virus, etc. All kinds of events regardless of their type make some changes in the environment. If we somehow manage to identify such changes in the early stages we can save many lives and prevent the potential damages. The early event detection systems are developed for such purposes. In these systems, multiple streams of pre-diagnostic health records \cite{shmueli2010statistical,henning2004overview} such as daily counts of doctor/hospital/emergency room visits, over-the-counter medication sales, work/school absences, animal illness or deaths, internet-based health inquiries are being monitored simultaneously to trace the event footprints.

\begin{figure}[ht]
 \begin{center}
	\includegraphics[width=\textwidth]{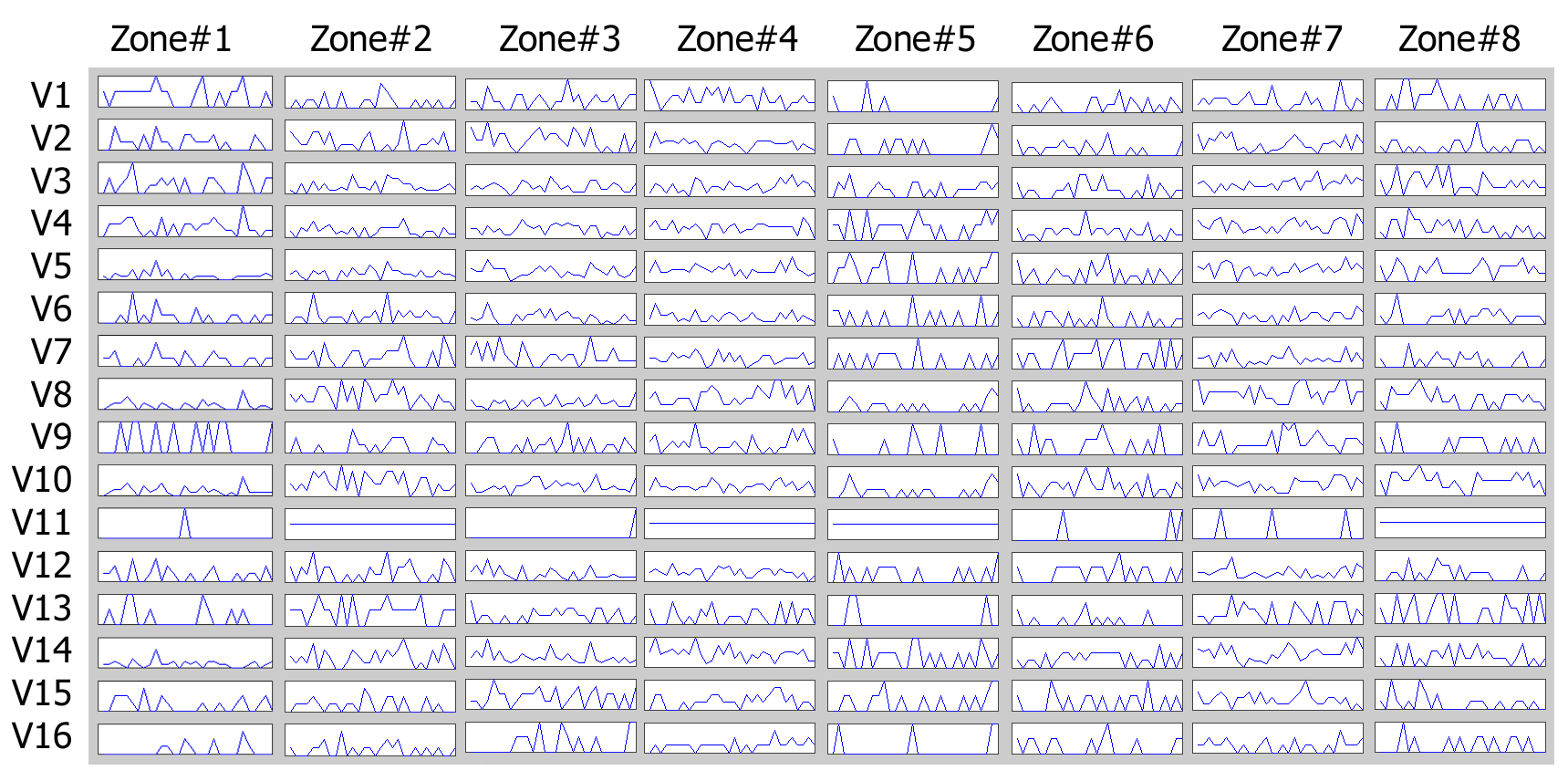}
 \end{center}
 \caption{A sample complex system in syndromic surveillance that generates 128 time series for 16 features and 8 spatial regions. } \label{fig:timeseries}
\end{figure}

Figure \ref{fig:timeseries} demonstrates an example of a complex data stream in synonymic surveillance systems. As it can be seen, this system measures 16 features aggregated daily within 8 different regions. Hence, the system generates 128 time series. Our goal is to monitor this complex system and signal an alarm when something strange occurs. One straightforward approach for monitoring such system is to monitor each individual time series and then apply an anomaly detection technique (e.g. Control chart) on each. This approach, however, imposes much higher false alarm rate. Because pre-diagnostic streams of indicators are weak and noisy signals \cite{cooper2004bayesian} and applying detectors on each individual signal results in multiple hypothesis testing problem \cite{WongMCW2005}. For instance, suppose that we reject null hypothesis when the $p-value \prec 0.05$, for a single hypothesis test, the probability of making a false discovery is equal to 0.05. Now assume that we do the test for each of 128 time series. Probability of false alarm could be as bad as: $1 - (1 - 0.05)^{128} = 1.00 >> 0.05$. 

Existing univariate methods include statistical process control based approaches \cite{williamson1999monitoring,hutwagner2003bioterrorism}; Time series analysis and signal processing based approaches, including singular spectrum analysis(SSA) \cite{moskvina2003algorithm}, Box-Jenkins models \cite{reis2003time}; Wavelet \cite{zhang2003detection}, Hidden Markov Model(HMM) \cite{rath2003automated,le1999monitoring}; and regression \cite{serfling1963methods}. The univariate methods since only monitor a single variable are not proper techniques for handling the complex data in synonymic surveillance. Besides, if we monitor each individual feature independently without taking into account the correlation between them, we then likely confuse the measurements error and noises with the events. 

The other category of methods is multivariate methods that are able to monitor multiple streams. These methods include Hotelling T2 \cite{ye_hotelling_2000}, multivariate CUSUM and EWMA \cite{macgregor1995statistical}, principal component analysis (PCA)\cite{Ku1995179}, multivariate HMM \cite{moore2006combining}, vector autoregression (VAR) \cite{Enders1993,barros2003intervention} and vector autoregression moving average (VARMA) \cite{Dong2010133}. There is also a sub-group of multivariate methods that operates on categorical data and looks for interesting rules via contrast set mining techniques. STUCCO \cite{bay1999detecting} and Emerging Patterns \cite{dong1999efficient} are instances of such techniques. Multivariate \textit{temporal} methods, despite of their wide application in many areas, are not well-suited to syndromic surveillance and outbreak detection problems where \textit{geographic} dimension is widely involved.

The methods that take into account geographic dimension are twofold: spatial and spatiotemporal. Spatial methods such as spatial scan statistics \cite{kulldorff1997spatial} do not capture the temporal fluctuations of the data and only operate on spatial data. Spatiotemporal methods instead take into account both spatial and temporal dimensions. Space-time scan statistics (STScan) is of this group that can operate both on univariate count data \cite{kulldorff1997spatial,kulldorff1995spatial,kulldorff1999spatial, kulldorff1998evaluating} and multivariate data \cite{kulldorff2007multivariate}. Univariate STScan is not adequate for syndromic surveillance for the same reason mentioned for univariate temporal methods. Multivariate STScan also has some drawbacks that make it be inappropriate for the introduced problem. On one hand they assume that the environment is static and do not consider seasonal effects and on the other hand they are developed for retrospective and offline analysis. Therefore, this group of techniques is not also suited to the problem.

There is another group of techniques such as PANDA \cite{Cooper2004} that use a causal Bayesian network to model spatiotemporal patterns of outbreaks. These methods not only explicitly compute the probability of events, but also are able to operate in real time settings through incremental updating of the Bayesian network. However, the main criticism against these techniques is that tuning the primary parameters requires a deep prior knowledge that is not available most of the time. Therefore, these methods are considered domain specific and their application has remained limited. 

Among many existing techniques and algorithms, the most suited approach to the introduced problem is WSARE \cite{WongMCW2003,WongMCW2005} that is able to handle multivariate data along spatial and temporal dimensions. WSARE searches for surprising rules in data streams given some baseline reference. The baseline creation strategy varies in different version of the algorithm. WSARE 2.0 uses raw historical data from selected days, WSARE 2.5 uses all historical data that match the environmental attributes and WSARE 3.0 models the baseline distribution using a Bayesian network. Opposed to PANDA where Bayesian network is created manually, WSARE 3.0 learns the Bayesian network from historical set. Therefore, is not as such domain-dependent as PANDA. WSARE has been successfully applied and merited in many real world problems such as in bioterrorism surveillance for 2002 Winter Olympics \cite{gesteland2003automated} and Israel influenza type B outbreak and Walterton outbreak \cite{kaufman2005using}. However, the main criticism about WSARE is its high rate of false alarms \cite{BuckeridgeBCHM2005}. WSARE opposed to other techniques, processes the data from bottom to up. Therefore, instead of overall changes in the whole data, it tracks the changes in subgroup of data. Therefore, it is sensitive to small changes and consequently presents lower detection delay, however, comes with more false alarm rate.

The methodological differences between our proposed method and WSARE are as follows. 1) WSARE is a bottom-up rule-based approach while our method is a middle approach between bottom-up and top-down that tracks both high level and dimension-based changes in the data subspace. 2) Our approach takes into account both multi-linear and multi-way correlations in data while WSARE is not able to capture such complexity; and 3) Our method is suitable only for alarming purposes and cannot explain about subgroup of the data that cause the alarm, while WSARE can be used for both purposes. 4) The statistical significance of the alarms in WSARE is computed via Monte Carlo simulation while in our approach is computed by statistical process control techniques.

In overall, the main objective in syndromic surveillance systems is to detect events in a timely manner before they turn into an epidemic. This early detection has important functions in both mortality saving and prevention of economic losses. An estimation by DARPA shows that a two-day improvement in detection time could reduce fatalities by a factor of six \cite{Neill2006}. Another study states that improvements of even an hour in detection can reduce the economic impact of by a hundred million of dollars \cite{Wagner01theemerging}. To reach this objective, any capable signal is required to be considered. However, this is somehow problematic, since involving more signals results in more false alarms. In the recent years the emphasis of the developed algorithms in syndromic surveillance has been focused more on the early detection and rate of false alarm is rarely taken into account. This is while the recent studies show that the false alarm rate can have an inverse effect as bad as delay in detection. A recent study concerning the warning system for tornado events \cite{simmons2009false} reveals that tornadoes occurring in the regions with a high false alarms ratio kill and injure more people. A statistically significant effect of false alarms is identified in this study: A one-standard-deviation increase in the false alarm ratio increases expected fatalities by between 12\% and 29\% and increases expected injuries by between 14\% and 32\%. 

\begin{figure}[ht]
 \begin{center}
	\includegraphics[width=0.5\textwidth]{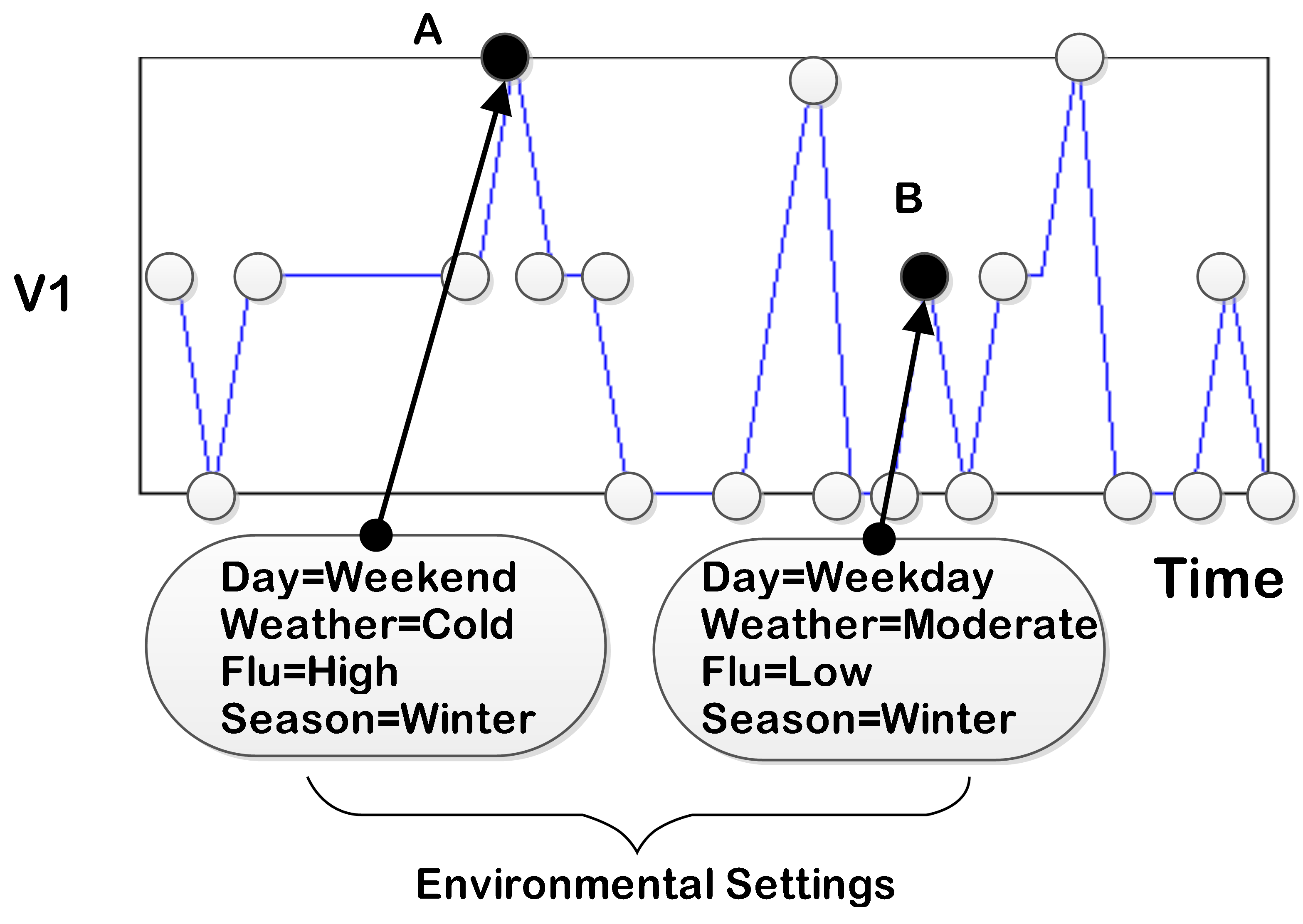}
 \end{center}
 \caption{The environmental setting affects the data items.} \label{fig:seasonal}
\end{figure}

Besides, opposed to anomaly or outlier detection problems, which it is assumed that the process occurs in an isolated and static environment in synonymic surveillance systems we deal with dynamic and time-changing environment. In such environments, attributes such as day of the week, holiday, weather, etc. affects the whole or part of the system behavior. Figure \ref{fig:seasonal} illustrates an individual time series corresponding to the feature $V1$ and $Zone 1$ in Figure \ref{fig:timeseries}. As it can be seen, in point $A$ due to cold weather and high rate of influenza rate, we have a higher count comparing point B. Such effects impose another kind of complexity to the event detection problem in syndromic surveillance which is required to be taken into account along with other issues. 

In this paper, we propose a novel event detection methodology that considers both data complexity and time-changing environmental issues in syndromic surveillance. The concentration of this work is to reduce the false alarm rate of early event detection systems. Our contributions are as follows.

\begin{itemize}
	\item  To the best of our knowledge this is the first time that the tensor decomposition techniques \cite{kolda2009tensor} is applied to the syndromic surveillance problem \textit{with space and time dimensions}.
	\item  We use the changes in data dimensions and data correlation structure as an effective criteria for event detection. 
	\item  We introduce a novel and effective approach for baseline data creation that can infer baseline for unseen environmental settings.
\end{itemize}

The rest of the paper is organized as follows. In section  \ref{sec:solution} we introduce the proposed solution and our developed algorithm EigenEvent. The section \ref{sec:evaluation} includes experimental evaluation, including the introduction of the data set, performance evaluation and sensitivity analysis. The last section concludes the exposition presenting the final remarks.

\section{Proposed method} \label{sec:solution}

\begin{figure}[ht]
 \begin{center}
	\includegraphics[width=10cm]{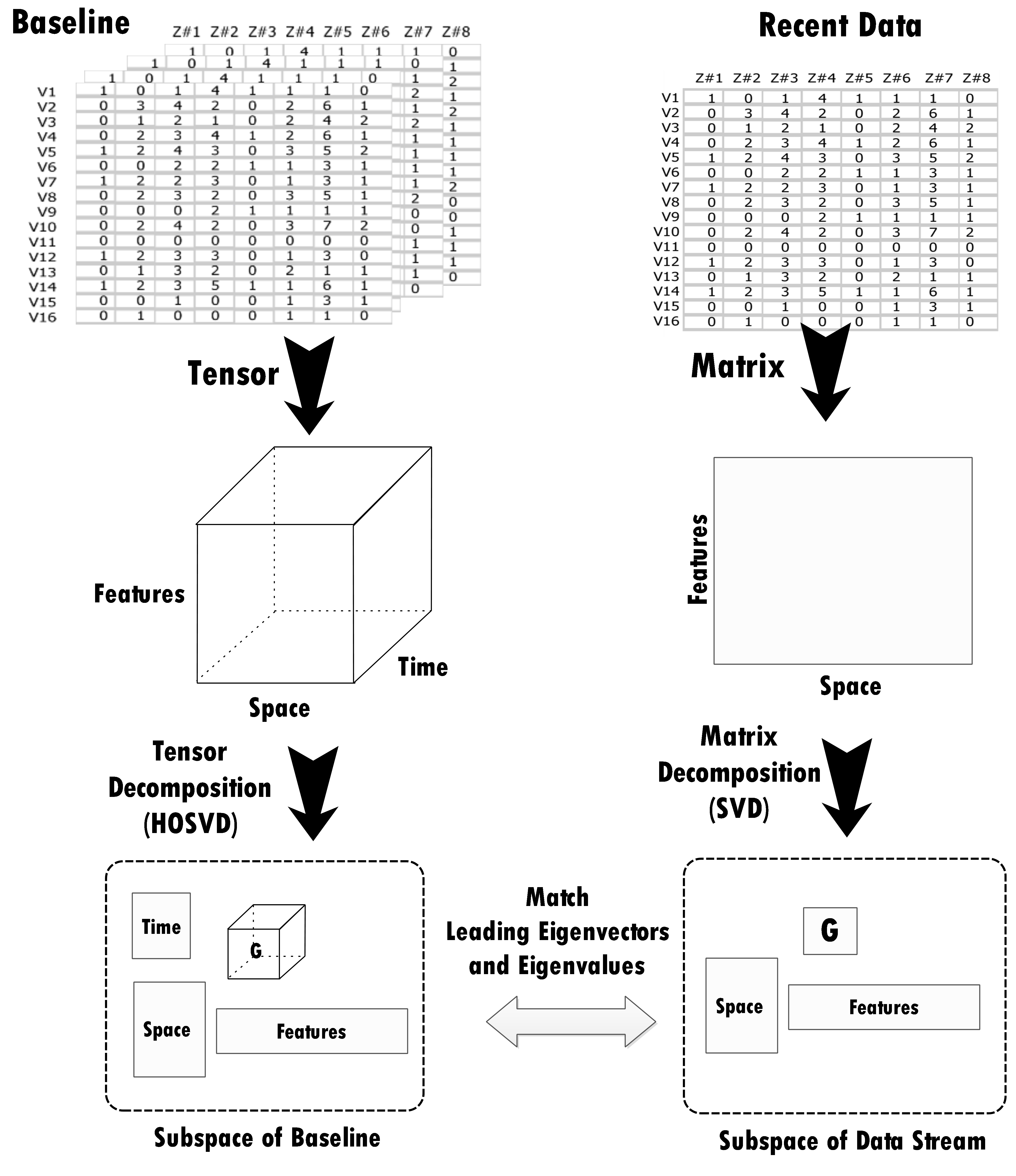}
 \end{center}
 \caption{Snapshot of the proposed solution at a hypothetical timestamp. We detect events through tracking changes in the subspaces of baseline and recent data.} \label{fig:proposed}
\end{figure}

\subsection{The idea} \label{sec:idea}

The fundamental idea that is used to develop the method relies on tracking changes in the subspace. This is impossible unless we could match the recent data with a \textit{baseline reference}. However, in streaming settings, data itself is time-changing due to the effect of the dynamic environment on the data items. Therefore, using a static baseline seems to be inappropriate for dynamic environments. We propose a dynamic baseline set creation strategy which takes into account both seasonality and non-stationarity. The main novelty of our method is that we not only track changes in the feature subspace, but in the subspace of other dimensions. 

Figure \ref{fig:proposed} demonstrates an illustrative example of our proposed method. Each day we receive a chunk from a complex data stream. In the data stream model this can be translated to the sliding window with fixed size of one day across the data stream. The window here is more complicated than a one-dimensional window in temporal data processing. Each window is a two-dimensional matrix of $Space \times Features$ (top-right matrix). Each cell in the matrix corresponds to the count of a feature in specific regions. With respect to the sliding window environmental setting, we generate a dynamic baseline tensor with order of $Space \times Features \times Time$ (top-left tensor) which is being fed from the historical data. This baseline tensor is built in each step or cycle of the algorithm run. The baseline tensor is composed of some previously arrived sliding windows that are combined in a particular order. We decompose the recent matrix and the baseline tensor to a lower-rank subspace and then match their pairwise eigenvectors and eigenvalues. We signal an alarm if we observe any unexpected difference in the match. 

\begin{figure}
 \begin{center}
	\includegraphics[width=6cm]{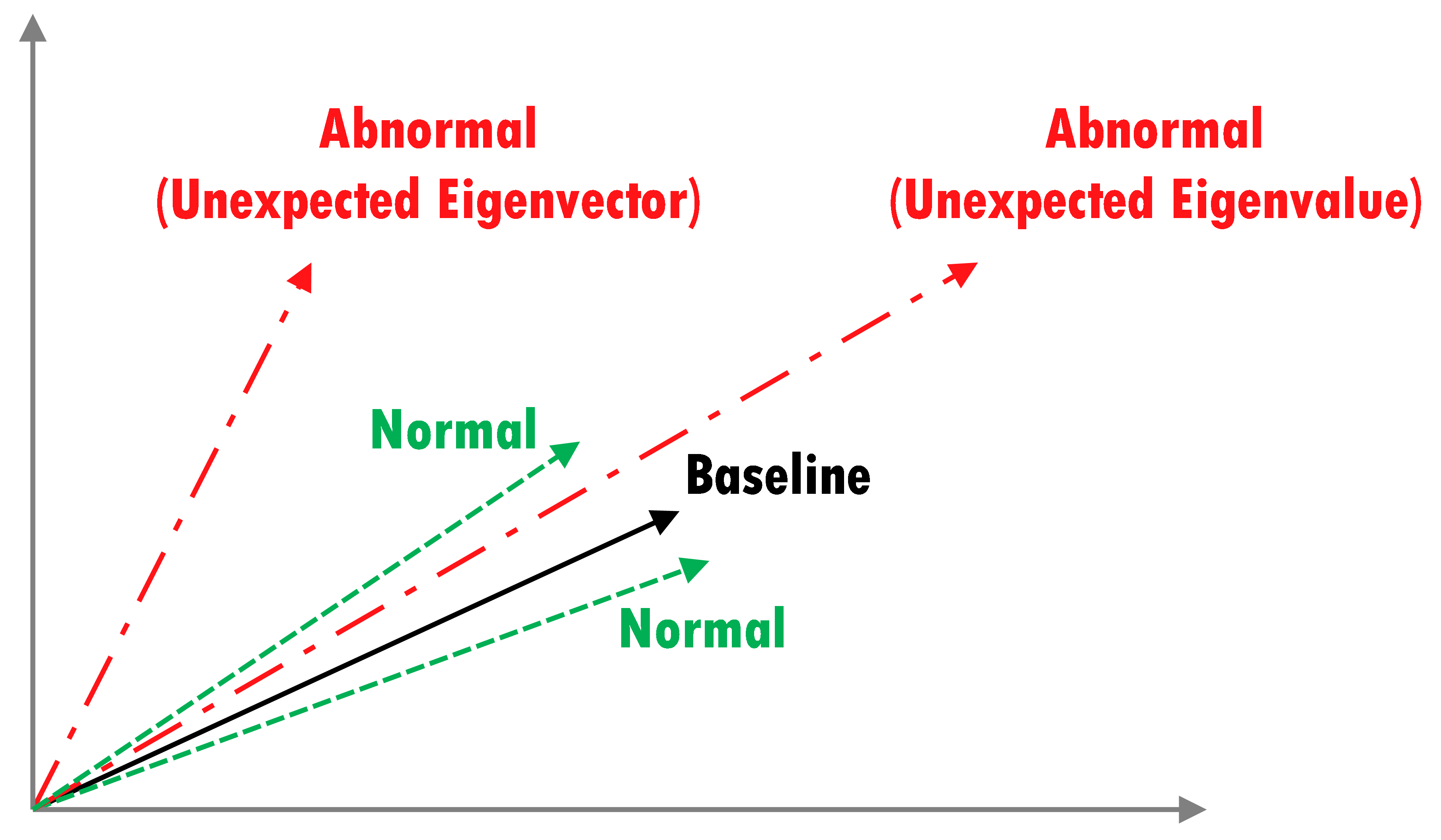}
 \end{center}
 \caption{A simplified example showing how events can be detected by tracking changes in the eigenspace. If the distance between the sliding window eigenvector and baseline eigenvector is higher than expected, then the window is marked as abnormal. Also, if the ratio of window eigenvalue to the baseline eigenvalue is higher than expected, the window is marked as abnormal as well.} \label{fig:Eigenspace}
\end{figure}

Figure \ref{fig:Eigenspace} illustrates the eigenspace of both baseline and recent matrix. The solid vector in this figure corresponds to the baseline tensor. The direction of this vector corresponds to the principal eigenvector corresponding to a dimension and the length of the vector corresponds to the principal eigenvalue. When we receive a matrix we decompose it to the eigenvectors and eigenvalues and then match the obtained principal eigenvalue and principal eigenvectors to the reference vector (solid vector). We signal an alarm if the matrix eigenvector has a considerable difference in direction (eigenvector) or length (eigenvalue). For instance, dashed lines in the figure correspond to those matrices that have close eigenvector to the baseline eigenvector and have the close eigenvalue (vector length). Such matrices are considered normal by EigenEvent. Dash-dot lines in the figure on the contrary are related to abnormal matrices that have an unexpected eigenvector (unexpected vector direction) or unexpected eigenvalue (unexpected vector length) with respect to the baseline.  

\subsection{Proposed Algorithm: EigenEvent} \label{sec:alg}

In this section we describe our proposed algorithm, which is called EigenEvent. As it is presented in Algorithm \ref{algorithm:EigenEvent}, the inputs are as follows: sliding window \textit{D} with length of one day; \textit{t} which is the sequence number; \textit{e} is a number corresponding to the environmental setting of the day. For instance, 
the environmental setting \textit{1214} is related to: day=weekend(1), weather=cold(2), flu=high(1), season=winter(4). 
The algorithm as a result outputs a p-value indicating the statistical significance of the sliding window. A very low p-value can be interpreted as an event signal.

\subsubsection{Data Processing and Decomposition} \label{sec:methodstep1}

The first phase is to transform the sliding window to the matrix format of $Space \times Feature$ (line \ref{line:CTensor}). To assess the abnormality of sliding window we need a baseline reference to match with. Two strategies can be utilized, one is to compare the window with the previous data and another strategy is to compare the window with previous data that have the same environmental setting. We use a combined strategy that takes into account both (see section \ref{sec:baselinetensor}) and produce the dynamic baseline set according the context corresponding to the window (line \ref{line:BTensor}). As a result, baseline is presented as a tensor of $Space \times Feature \times Time$. We then apply SVD \cite{klema1980singular} on window matrix and higher order SVD (HOSVD) \cite{kolda2009tensor,de2000multilinear} on the baseline tensor and for each dimension we take the principal eigenvector and eigenvalue.

Note that EigenEvent does not concern about the feature selection (selection of pre-diagnostic signals that are required to be monitored). Feature selection, however, may be performed via standard feature selection techniques or via domain experts or a combined technique. Nevertheless, feature selection is one the most important steps in a data mining process that is required to be taken into account. Selection of inappropriate signals may result in higher false alarm or more detection delay. The well-known over-fitting problem may happen here as well. Leinweber in an article entitle \textit{stupid data miner tricks: overfitting the S\&P 500} names some of such problems. He finds a strong correlation between butter production in Bangladesh and S\&P 500 (stock market index) over a ten year period. This implies that the selection of appropriate signals still is human-dependent and cannot be fully automated.

\subsubsection{Subspace Matching} \label{sec:methodstep2}

The next phase is the matching phase. If we denote the principal eigenvalue of baseline with $\lambda_b$, 
the principal eigenvalue of window with $\lambda_s$, the principal eigenvector of baseline with $X_b$ and the principal eigenvector of window with $X_s$, we can define the ratio of eigenvalues and Euclidean distance of eigenvectors respectively as:

\begin{equation}
d_{1,t}=\frac{\lambda_s}{\lambda_b} 
\end{equation}
\begin{equation}
\|d_{2,t}\| = (X_s,X_b).
\end{equation}

We keep the historical distances in two vectors of \textit{$vd_1$} and \textit{$vd_2$} for eigenvalues and eigenvectors respectively, such that at time \textsl{t} we have $vd_1=(d_{1,1}, d_{1,2}, ..., d_{1,t-1})$ and $vd_2=(d_{2,1}, d_{2,2}, ..., d_{2,t-1})$.  Having $d_{1,t}$, $d_{2,t}$, $vd_1$ and $vd_2$ we can compute the z-scores corresponding $d_{1,t}$ and $d_{2,t}$ as follows.

\begin{equation}
z1=\frac{d_{1,t}-\mu_{vd_1} }{\sigma_{vd_1} }
\end{equation}
\begin{equation}
z2=\frac{d_{2,t}-\mu_{vd_2} }{\sigma_{vd_2} }
\end{equation}

Where $\mu_{vd_1}$ and $\mu_{vd_2}$ denotes the mean and $\sigma_{vd_1}$ and $\sigma_{vd_2}$ denote standard deviation of vector $vd_1$ and $vd_2$ respectively. 

Although z-scores alone can be used along with a threshold for alarming purpose, since most related event detection algorithms in the literature outputs p-value, we may want to transform z-scores to the corresponding p-value to ease the comparison task. We can use the following equation to derive the p-value from the z-score:

\begin{equation}
P(z)=\frac{1}{\sqrt 2\pi }\int_{-\infty}^{z}e^{\frac{-t^2}{2}}dt
\end{equation}

\subsubsection{Indicator Selection} \label{sec:methodstep3}

As we already explained, HOSVD and SVD decompose the complex data into smaller subspaces (eigenspace). Tensor and matrix decomposition methods are robust against the noises. However, in the case that we have some missing values we need to use specific types of SVD \cite{kurucz2007methods}.

We have three elements in the eigenspace that can be matched: principal eigenvector of spatial and feature dimensions and the principal eigenvalue. We may observe three kinds of changes in the match. The first kind includes an overall change in the system which is more related to the late days of outbreak period when we have both infection and outbreak. This kind of event must be reflected in a significant change in the ratio of eigenvalues ($d_{1,t}$). The second kind of change occurs when an event agent (e.g. Virus) begins to spread over the geographical space. This type of event also is reflected in the changes in the spatial eigenvector pairwise distance ($d_{2,t}$). The third kind is the change in the feature values. This event type can be reflected in the eigenvectors corresponding to feature dimension. However, as we show later due to the noisy properties of the feature dimension, this kind of indicator is not such helpful.  

We propose a new strategy that is able to detect both overall and dimension-based changes in the system. We monitor the system using a combination of indicators, including  eigenvalue and different eigenvectors and the compute the p-value corresponding to each combination for each sliding window. Then we take the minimum p-value as the algorithm output (line \ref{line:Pvalue}). Suppose that we have three p-values of 0.01, 0.12 and 0.43 corresponding to the pairwise match between the principal spatial eigenvector, the principal feature eigenvector and the principal eigenvalue respectively. The EigenEvent algorithm reports the minimum p-value (0.01) as the output. These above mentioned p-values indicate three facts about the system: 1) No overall change has occurred in the system, because p-value corresponding Eigenvalue is considerably high; 2) No significant change is occurring in the feature values; 3) A significant change is occurring in the spatial dimension. We may infer that data items despite of showing normal behavior in the features are showing different behavior in geographical space and hence we probably are in the outbreak phase. The minimum p-value selection strategy lets us to detect all above kinds of changes and subsequently makes the algorithm sensitive to changes in both overall system behavior and the dimension level.

\begin{algorithm}
\caption{EigenEvent} \label{algorithm:EigenEvent}
\begin{algorithmic}[1]
\Statex //D: Recent data (Right top table in figure \ref{fig:proposed})
\Statex //C: Recent data in the format of matrix $Space \times Features$
\Statex //t: instant (e.g. t=3 means after 3 days of monitoring started)
\Statex //e: Recent data Env. Setting (e.g. 1214=\{day=weekend, weather=cold, flu=high, season=winter\})
\Statex //EV: Environmental setting vector (e.g. [1214,1321,3214,1456])
\Statex //H: Historical Tensor
\Statex //vd1: vector of principal Eigenvalue distances ($d_{1,1}, d_{1,2}, ..., d_{1,t-1}$) 
\Statex //vd2: vector of principal spatial Eigenvector distances ($d_{2,1}, d_{2,2}, ..., d_{2,t-1}$) 
\Statex //B: Current generated Baseline tensor (Tensor $Space \times Time \times Features$ in Figure \ref{fig:proposed})
\Statex //P-value: Statistical Significance of the recent data (e.g.  Signal an alarm when $p-value \prec 0.05$)

\Require D, t, e
\Ensure P-value
\State $Matrix\ C \gets D$  \label{line:CTensor}
\State $Tensor\ B \gets BaselineTensorUpdate (B,H, t, e, EV, C)$ \label{line:BTensor}
\State HOSVD(B): $X_b \gets principal\ spatial\ Eigenvector$, $\lambda_b \gets \ principal\ Eigenvalue$ \label{line:BDecompose}
\State SVD(C): $X_c \gets principal\ spatial\ Eigenvector$, $\lambda_c \gets \ principal\ Eigenvalue$ \label{line:CDecompose}
\State $d_1=\frac{\lambda_b}{\lambda_c}$ 
\State $\|d_2\| = (X_c,X_b).$ 
\State p1= p-value of $d_1$ given $vd_1$
\State p2= p-value of $d_2$ given $vd_2$
\State $P-value \gets Min[p1, p2]$ \label{line:Pvalue}

\If e exists in EV \label{line:Ifstart}  
\State $vd_1 \stackrel{add}{\longleftarrow}  d1$  \label{line:Ifc1}
\State $vd_2 \stackrel{add}{\longleftarrow} d_2$ \label{line:Ifc2}\label{line:Ifend}
\EndIf 

\State $H \stackrel{add}{\longleftarrow}  C$
\State $EV\stackrel{add}{\longleftarrow}  e$
\Statex
\Function{BaselineTensorUpdate}{B,H,t,e,EV,C}\
\If{ B is empty} $B\stackrel{add}{\longleftarrow} C$  \Else
\State k=0
\For {i=1 to t-1}
\If {EV(i) == e}
\State $k=k+1$
\State $B(k) \gets H(i)$
\EndIf    
\EndFor  
\EndIf
\State Return B
\EndFunction
\end{algorithmic}
\end{algorithm}

\subsubsection{Dynamic Baseline Tensor} \label{sec:baselinetensor}

There should be a criterion to estimate the abnormality of the recent data. As is mentioned before, two types of common criteria includes comparing with the previous data and comparing with only the previous data that match the current environment settings. Both of these criteria are vulnerable. The first criteria fails when data contains seasonal effects and second one fails when there is no enough historical data matching the recent environmental setting. To solve this problem a typical inference usually is performed, for instance, a causal Bayesian network is constructed in WSARE 3.0 \cite{WongMCW2003,WongMCW2005} so that when there is no enough historical data, baseline is inferred from the constructed Bayesian network. This approach, however, only make inference about the days their corresponding environmental settings cannot be found in the baseline set. In the rest of the time it compares the recent window with the previous data that match the current environment settings. This approach can be vulnerable as well, since the correlation of the current window with the recent data is ignored. We introduce another way of baseline set selection which is a combination of both ideas. We assume that the recent data is not only related to the previous data and data with the same environmental settings, but also to data with the most repeated environmental settings. In fact, our baseline tensor is a combination of previous data, data with the same environmental setting and data from most frequent environmental settings. The main advantage of this approach is that it does not fail when deal with an unseen environmental setting.

\begin{figure}
 \begin{center}
	\includegraphics[width=12cm]{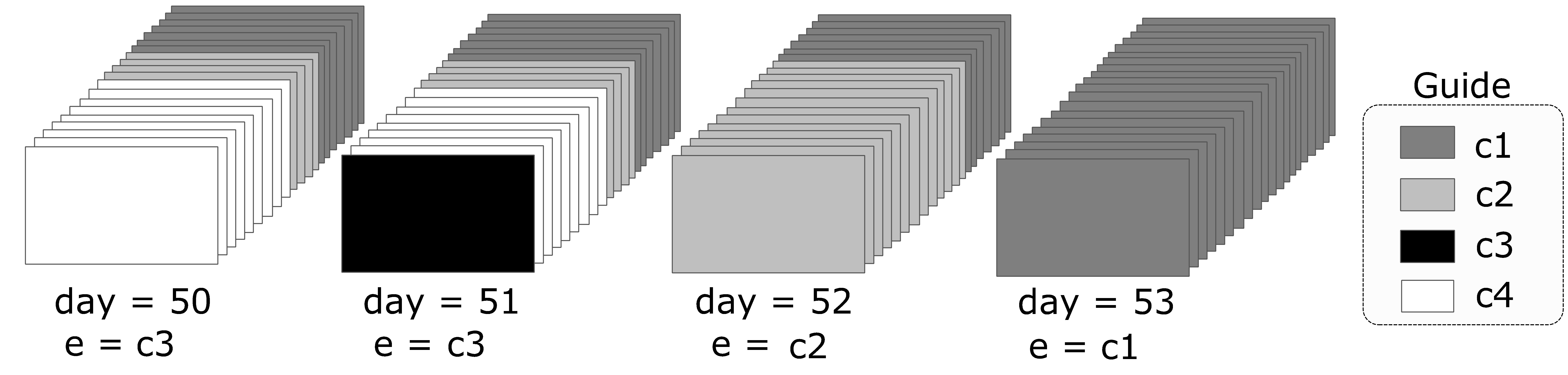}
 \end{center}
 \caption{A Sample of dynamic baseline tensor creation process between day 50 to day 53. Each plate represents a daily matrix of $Space \times Features$ in historical set. The dynamic baseline tensor is a combination of such matrices in a particular order. \textit{e} also denotes the environmental setting of the day.} \label{fig:pop}
\end{figure}

The function \textit{BaselineTensorUpdate} in Algorithm \ref{algorithm:EigenEvent} receives six inputs, including $B$ (current baseline tensor); $H$ (whole data, historical tensor); $t$ (instant number); $e$ (the recent environmental setting); $EV$ (vector of all environmental settings seen yet); and $C$ (recent matrix) and outputs the updated baseline tensor  $B$. It first checks that whether the tensor $B$ is empty. In the case that  $B$ is empty, $C$ is added to  $B$. 
Then we search in historical tensor $H$ for data that match the recent environmental setting. Next, it rewrites the first \textit{k} matrices of tensor $B$ with the matched items.

An illustrative example of the procedure is demonstrated in Figure \ref{fig:pop}. The figure is a snapshot of the system at four hypothetical days between days 50 to 53. From the figure we also can observe four distinct environmental settings, which are shown with different colors and that their corresponding name is demonstrated in the guide table. Each cube in the figure represents a baseline tensor and each plate inside the cubes is a $Space \times Features$ matrix from the historical set. At day 50 the baseline tensor is composed of 20 matrices such that 9 of matrices are from setting {\bf c4}, 
4 matrices from setting {\bf c2} and 7 matrices from setting {\bf c1}. 
We also assume that the context {\bf c1} is the dominant environmental setting with 20 times occurrence. 
The dominant context is the most frequent setting in all the history. 
For this reason, all baseline tensors, in Figure \ref{fig:pop}, include 20 matrices, given that the length of the baseline tensor is equal to the number of occurrences of the dominant context. 

Now let's explain how a dynamic baseline set is generated. At day 50, we receive a matrix with setting {\bf c1}. We search in historical tensor $H$ for a match with {\bf c1} setting, but we do not find, so the function \textit{BaselineTensorUpdate} returns input $B$ unchanged. On day 51, we again receive a matrix with the setting {\bf c1}. We again search for a match in $H$. This time we find one match, because one day before (day 50) the setting has been {\bf c1}. Therefore, we rewrite the first $k$ elements of $B$ Tensor with $k$ found matrices. In this case since we find only one match, $k$ is equal to 1. At day 52 we receive a matrix corresponding with environmental setting \textbf{c2}. We search in H for a match and suppose that we find 13 matrices. Hence, $k$ will be equal to 13, so we rewrite the first 13 elements of the baseline tensor with the matched 13 matrices. As it can be observed at day 52, setting \textbf{c2} has been the dominant setting versus \textbf{c3} and \textbf{c4} settings, however, still \textbf{c1} dominates \textbf{c2} (\textbf{c1} setting has more repeats comparing \textbf{c2}), therefore, the baseline tensor is composed of matrices with most dominant settings with preference to the recent data. Finally, on day 53, we receive a matrix with setting \textbf{c1}. We search in $H$ for a match and we find 20 matrices ($k=20$), thus we rewrite first 20 elements of the baseline tensor with matrices corresponding \textbf{c1} settings. At this moment, the whole baseline tensor is filled with only matrices with setting \textbf{c1}. This procedure repeats and repeats. However, the size of baseline tensor always stays fixed to the repeat count of the most repeated environmental settings.

\subsubsection{Updating Step} \label{sec:methodstep3}

In this step we update the vector of distances (line \ref{line:Ifc1}-\ref{line:Ifc2}). 
We add the distances to the vectors if their corresponding contexts has been already seen. 
If we have a matrix with an unseen environmental setting, we do not add the computed distance to the vector of distances. Because an inference for this setting is approximate and adding the distance obtained from this approximation is not adequate for keeping. We finally update historical tensor and vector of environmental settings.

\section{Evaluation} \label{sec:evaluation}

\subsection{Data set} \label{sec:dataset}

Validation of event detection algorithms is basically a difficult task due to the type of required data \cite{WongMCW2005,BuckeridgeBCHM2005,ShmueliGFS2006}. To evaluate the algorithms, the event occurrence period is required to clearly be labeled in the data. This requires a knowledge expert to look into the data and specify the event period manually, making this task infeasible. Benchmark data sets that are already used for change detection and anomaly detection are not appropriate for our research purpose, because, on one hand, most of the time they do not have seasonality property and on the other hand do not contain multi-way property. We recently \cite{fanaeegama2013prai} proposed a semi-automatic for labeling events in unlabeled data which is based on ensemble detectors and background knowledge from web. However, this approach also needs to have access to some sort of background knowledge which is not available in this domain. 

We use a benchmark data set used in \cite{WongMCW2003} including 100 data sets of a simulated disease outbreak. These data sets are generated using a Bayesian network simulator namely CityBN which generates temporal fluctuations based on a variety of factors such as weather and food conditions \cite{WongMCW2005}. The structure and parameter of this Bayesian network are manually adjusted. As is mentioned by the authors, this simulator produces extremely noisy data sets that are a challenge for any detection algorithm. This data set is publicly available online in \cite{WengM2013}. 

Table \ref{tab:datarecords} shows the characteristic of the original data sets. As it can be seen, this data is multi-way. It contains two dimensions of space and time and multiple variables. It also contains seasonal effects, because features are under influence of some environmental settings. Cardinality of each attribute is also specified in the table. As it can be seen, we have 9 distinct spatial regions and 730 temporal instants (days). We also have 16 (3+2+3+4+4) distinct time series and $4 \times 3 \times 2 \times 4$ possible environmental settings.

\begin{table}
\begin{center}
\caption{Characteristic of CityBN Data sets}\label{tab:datarecords}
		\begin{tabular}{|c|c|c|c|c|}
    \hline
    Field & Type& Cardinality & Sample Record\#1  & Sample Record\#2  \\ \hline
		XY & Spatial& 9& SW  & NE \\ \hline
    Daynum & Temporal& 730& 73779  & 74508 \\ \hline
    Age & Feature & 3& senior  & child \\ \hline
    Gender & Feature& 2& female  & male\\ \hline
		Action & Feature& 3& purchase  & evisit \\ \hline
    Reported symptom & Feature& 4& nausea  & respiratory \\ \hline
    Drug & Feature&  4& nyquil  & vomit-b-gone\\ \hline
    Flu Level in season & Environmental &  4& high  & decline  \\ \hline
    Day of week & Environmental& 3& weekday  & sat \\ \hline
    Weather & Environmental& 2& cold  & hot\\ \hline
    Season & Environmental& 4& winter  & sumer\\ \hline
    \end{tabular}
\end{center}
\end{table}

\subsection{Performance} \label{sec:exp}

Receiver operating characteristic (ROC) curve \cite{hanely1982meaning} measures the trade-off between sensitivity and specificity. ROC curve is widely used method for evaluation of anomaly detection and classification methods. However, ROC curve, even though summarizes the overall ability of the algorithm, does not evaluate the timeliness of detection which is critical in syndromic surveillance. An algorithm with the lowest false positive and the highest true positive rate that detect outbreaks with heavy delay is inappropriate for syndromic surveillance applications. In fact a system with this characteristic is more helpful for retrospective applications than the prospective applications like what is required in syndromic surveillance. One of the proper metrics for evaluation of  algorithms is Activity Monitoring Operating Characteristic (AMOC) curve \cite{FawcettF99} that evaluates the trade-off between specificity (false alarms) and timeliness (detection time). AMOC curve is widely used for evaluation of methods in syndromic surveillance \cite{cooper2004bayesian,WongMCW2005,siegrist2004bio,jiang2010bayesian}. Therefore, in this work we use AMOC curve for evaluation of our algorithm.

\begin{figure}
 \begin{center}
  	\includegraphics[width=10cm]{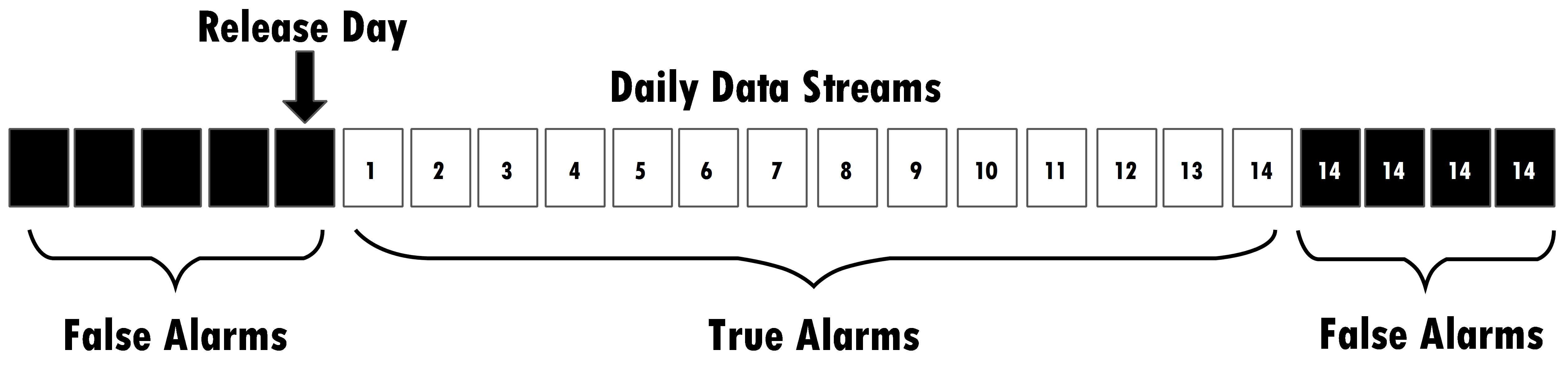}
 \end{center}
 \caption{Evaluation Strategy: Alarms in days with black color represents false alarms and in white days represents true alarms. Detection delay is also specified for each day inside the plates} \label{fig:delay}
\end{figure}

We use the same evaluation strategy as \cite{WongMCW2005}. Assume that the agent release occurs at timestamp $t$. A true alarm corresponds to a case where the alarm is raised in a period between $t+1$ and $t+14$. The alarms before or after this period are considered false positives. The detection delay is also defined as the temporal difference between the first alarm in the above period and the release time. In reality, the data of each day is processed tomorrow of that. Therefore, is not possible to detect event on the day of release. Thus, the optimum detection is tomorrow of the release (detection delay=1). This one day delay is also considered in CityBN simulation. Figure \ref{fig:delay} demonstrates that how we define false alarms and detection delay. If we signal an alarm in a period of 14 days after release it is marked as true alarm and if we signal an alarm before or after this period, it is marked as a false alarm. Detection delay is also specified in the figure as numbers in the plates. If we signal an alarm tomorrow of the release, we get only one-day delay which is the optimum condition. For any alarm after this period we define detection delay equal to 14 (as \cite{WongMCW2005}).

\begin{figure}
 \begin{center}
  	\includegraphics[width=8cm]{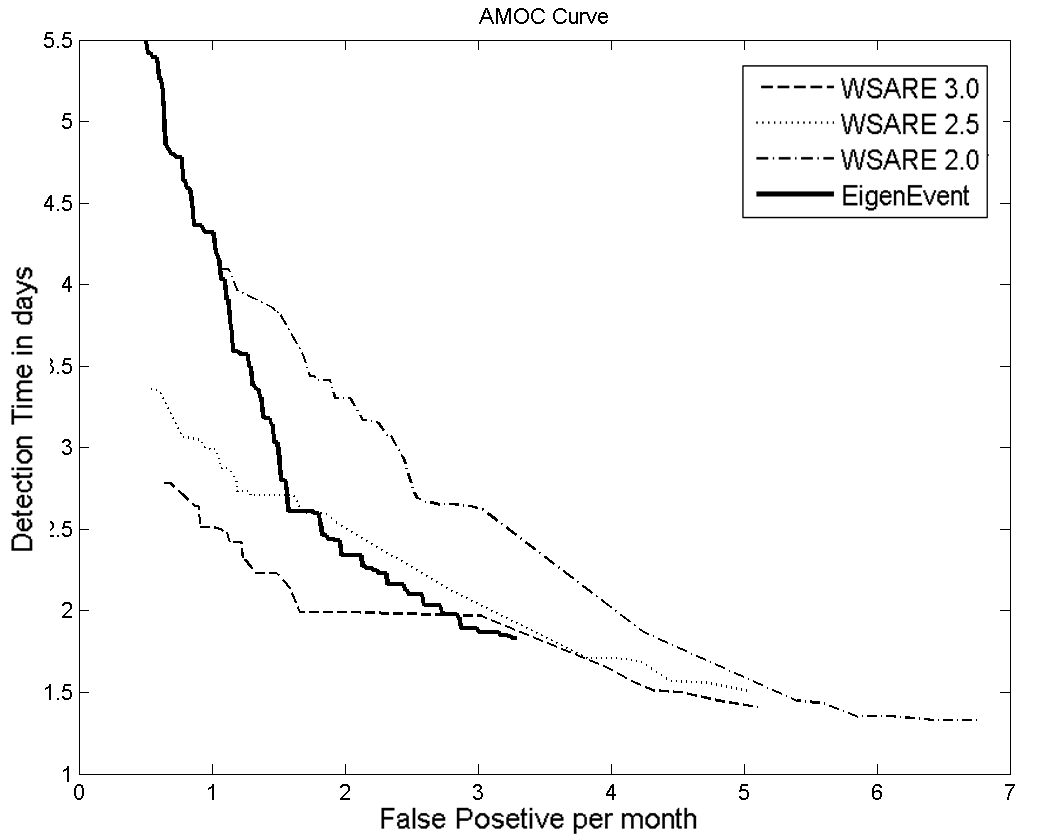}
 \end{center}
 \caption{AMOC Curve for EigenEvent vs. WSARE} \label{fig:amoc1}
\end{figure}

The outputs of both WSARE and EigenEvent are p-values indicating the statistical significance of recent data. Depending on the desired confidence level, we may signal an alarm. For instance, given a threshold as 0.05 we signal an alarm if the p-value corresponding to the recent data goes lower than 0.05. To assess the algorithms performances we use variable p-value threshold from 0.020 to 0.250 with the step of 0.001 (totally 231 p-values). Each data set has temporal size of 730 days. We use the first 365 days for training the primary baseline and the next 365 days for evaluation of the algorithms. Baseline set is also incrementally updated whenever a new window arrives after day 365. Note that agent release in all 100 data sets occurs in the second year and is guaranteed that the first year do not contain any release. A sliding window moves across the data from day 366 to day 730 and match each window with the baseline. If the match outputs a p-value below the threshold, then an alarm is raised. After we reached to day 730, we compute the number of false alarms and detection delays. We finally average the detection delay and false alarms for all 100 data sets and plot the AMOC Curve. In the AMOC curve, the x-axis indicates the number of false alarms per month and the y-axis measures the detection time in days. The optimal detection is one day detection delay with zero false alarm. The closer to the point (0,1) the better detection algorithm is. 

The results are shown in Figure \ref{fig:amoc1}. Although the curve corresponding EigenEvent seems different comparing WSARE, if we rotate the AMOC curve 90 degrees anticlockwise we observe the same pattern similar to WSARE 3.0. The difference is that EigenEvent performs better in terms of false alarm rate and performs worse in terms of detection delay. The intersection between the curves makes the overall comparison difficult. For instance, in a desired false positive rate from 2.8 to 3.3, EigenEvent is the best method both in terms of false alarm rate and detection delay. Nevertheless, to specify which of the algorithms are better in overall we need to compute the area under the AMOC curve \cite{QueT2008}, average delay and average false positive rate (see Table \ref{tab:comparison}). Obtained area under AMOC curve implies that EigenEvent outperforms all versions of WSARE. Its average false positive is considerably lower than all versions of WSARE. However, in terms of detection delay as was expected presents one more day delay. To have a separate look on both numbers of false alarms and detection delay, we also compute the average false alarms and detection delay for 231 p-values (from 0.020 to 0.250 with the step of 0.001). The results are presented in table \ref {tab:falsereates} and \ref{tab:delays} respectively. As it can be seen from the first table, EigenEvent in terms of false alarms, beats other methods in the majority of data sets. Regarding the detection delay even though is not the best, has detected events tomorrow of release in half of the data sets. 

The main reason for the differences in the performance is related to the methodological differences between EigenEvent and WSARE. EigenEvent opposed to WSARE is not a bottom up approach and subsequently is less sensitive to the small-scale changes and subsequently, presents less false positive rate. EigenEvent due to its less sensitivity to the small-scale changes reacts slower to the events. However, EigenEvent have this ability to track changes in the dimensions, for this reason does not suffer from the high false alarm rate problem of bottom-up approaches and heavy delay problem of the top-bottom approaches.

\begin{table}
\begin{center}
\caption{False positive rate (per month), Detection delay (in days), Area under AMOC Curve and Runtime (in seconds) Averaged for 100 data sets of CityBN}\label{tab:comparison}
\begin{tabular}{|c|c|c|c|c|}
\hline
Method&False positive&Detection delay&AUAMOC\\ \hline
WSARE 2.0&4.052439&2.163983&12.859000  \\ \hline
WSARE 2.5&2.739062&2.192338&9.885192 \\ \hline
WSARE 3.0&2.877031&\textbf{1.929134}&8.648379 \\ \hline
EigenEvent&\textbf{1.866439}&2.839827&\bf{8.027842}\\ \hline

\end{tabular}
\end{center}
\end{table}

\subsection{Runtime } \label{sec:exp}

\begin{table}
\begin{center}
\caption{Runtime (in seconds) Averaged for 100 data sets}\label{tab:runtime}
\begin{tabular}{|c|c|c|c|c|}
\hline
Method&Runtime\\ \hline
WSARE 2.0&59.2  \\ \hline
WSARE 2.5&105.3 \\ \hline
WSARE 3.0&838.4 \\ \hline
EigenEvent&\bf{16.8 }\\ \hline
\end{tabular}
\end{center}
\end{table}

Since in syndromic surveillance systems, data are often required to be processed in daily scale, computational efficiency receives less attention. In the unlikely case where data size becomes very huge and processing of data requires run-time of more than 24 hours (the process scale) then we have to come up with computational efficiency issues. Although, computational efficiency is not the claim in this research work, runtime in Table \ref{tab:runtime} indicates the superiority of EigenEvent over all versions of WSARE. EigenEvent requires only 16.8s to deliver the result. This is three times faster than WSARE 2.0, 6 times faster than WSARE 2.5 and 50 times faster than WSARE 3.0. The majority of this difference is related to two factors; WSARE exhaustively search the whole space while EigenEvent only tracks the changes in the correlation structure. The second factor is related to the method the approaches compute the p-value of alarms. WSARE exploits Monte Carlo simulations for computing the p-value while EigenEvent computes the p-value using statistical process control techniques which is lighter. 

In each time step, Eigenevent requires to perform a tensor decomposition and a matrix decomposition. The offline tensor decomposition (OTA) \cite{Sun2008} of the baseline tensor requires $O(T\prod_{i=1}^{M}n_i)$ where T is the temporal size of the tensor, M is the order of the tensor which in our case is equal to 3 (three dimensions of space, features and time) and $n_i (1 \leq i \leq M)$ is the dimensionality of the \textit{i}th mode (reshaped matrix in dimension \textit{i}). The matrix decomposition of the recent data also requires $O(N^2)$ for one-rank matrix decomposition. Therefore, in each step we require $O(T\prod_{i=1}^{M}n_i)$ + $O(N^2)$. Some approximation techniques are developed for reducing the computation time of the first term. For instance, \cite{Sun2008} proposed three different techniques including dynamic tensor analysis (DTA), streaming tensor analysis (STA) and window-based tensor analysis (WTA) that perform tensor decomposition more efficiently with much lower computation time. For instance, DTA requires computation time of $2 \sum_{i=1}^{M}r_in_i^2+\sum_{i=1}^{M}n_i\sum_{j=1}^{M}n_j$  where $r_i$ is the core size for each mode which in our case is equal to 1. Therefore, assuming the tensor with three dimensions (as our case study) we require only $2 \sum_{i=1}^{3}n_i^2+\sum_{i=1}^{3}n_i\sum_{j=1}^{3}n_j$ which is a tremendous improvement over OTA. For low-order tensor (i.e. $M \leq 5$) as is pointed out in \cite{Sun2008}, the diagonalization which is the main cost can be performed via faster approximation approaches. However, since computation efficiency is not the main concern in this part of our research we do not test all the available techniques. 

\subsection{Leading Indicators} \label{sec:combinedp}
\label{tab:indicator}

\begin{figure}
 \begin{center}
	\includegraphics[width=8cm]{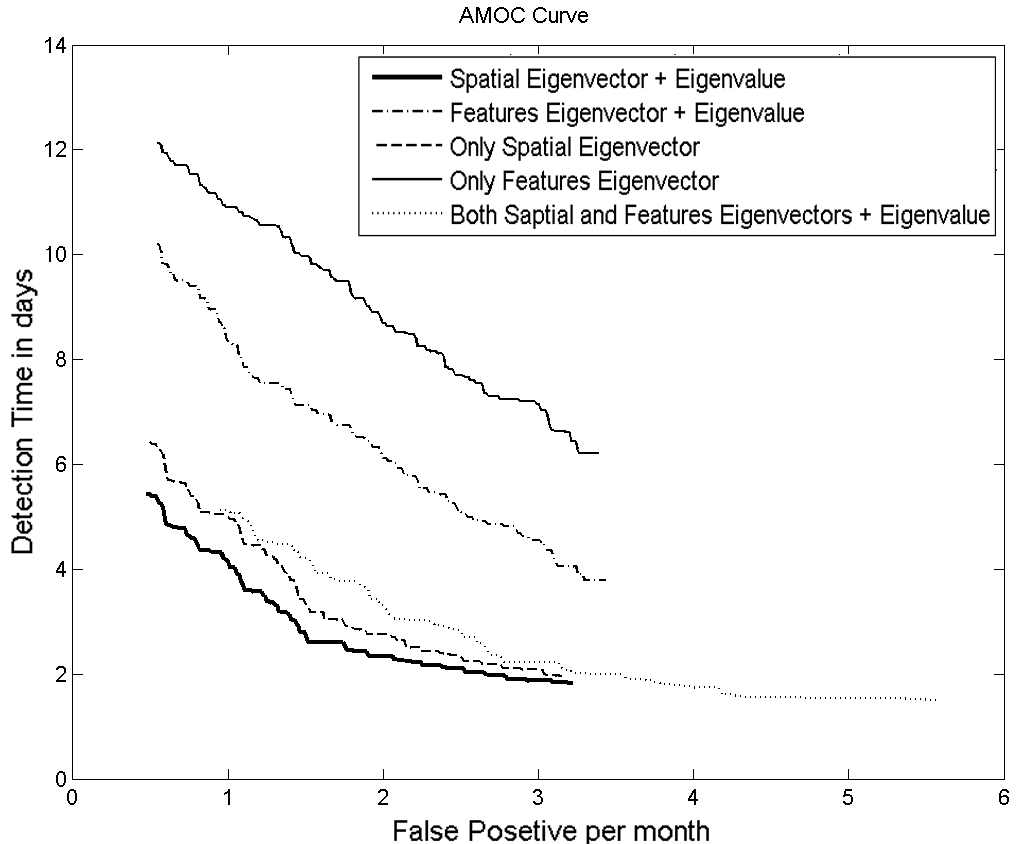}
 \end{center}
 \caption{Effect of indicators on the performance} \label{fig:indicator}
\end{figure}

As we mentioned before, EigenEvent algorithm tracks the deviation of sliding window eigenspace from the baseline tensor eigenspace for change detection. Now the question is that what elements of the eigenspace we should take for the match. Should we opt for eigenvectors corresponding to the spatial dimension or to the feature dimension. Should eigenvalue be used along eigenvector or eigenvector alone is enough. We examine five circumstances: the first condition is the default setting in the Algorithm (The optimum selection in line \ref{line:Pvalue} of Algorithm \ref{algorithm:EigenEvent}), and the rest are different combination of eigenvectors and eigenvalues. Figure \ref{fig:indicator} illustrates the AMOC curve for these different combinations. As it can be seen, by using only spatial eigenvector (without considering eigenvalue) we experience the same result but with more half-day average delay. In fact, involving of Eigenvalue in the change detection process provides earlier detection. We also study a condition where whole eigenspace is used. In this case we take into account both spatial and features eigenvectors along the eigenvalue. This leads to half-day delay earlier detection, but with 1.5 more false alarms. Excluding spatial eigenvector from the eigenspace matching also leads to lower performance both in terms of delay and false alarms. This result reveals that how the spatial dimension is important. In fact, temporal methods that exclude the spatial dimension loose lots of information. The reason is that feature signals are very noisy and detection of pattern of such noisy data comes with high false discovery. Instead, the spatial dimension is more stable and tracking changes in this dimension can be a better indicator for tracking particular events such as disease outbreaks (our case study), because, one of the key signatures of disease outbreak is movement in the space. This movement changes the constant patterns in the spatial dimension and subsequently this appears in the principal spatial eigenvector.

\subsection{Baseline Selection} \label{sec:baselineselection}

\begin{figure}
 \begin{center}
	\includegraphics[width=8cm]{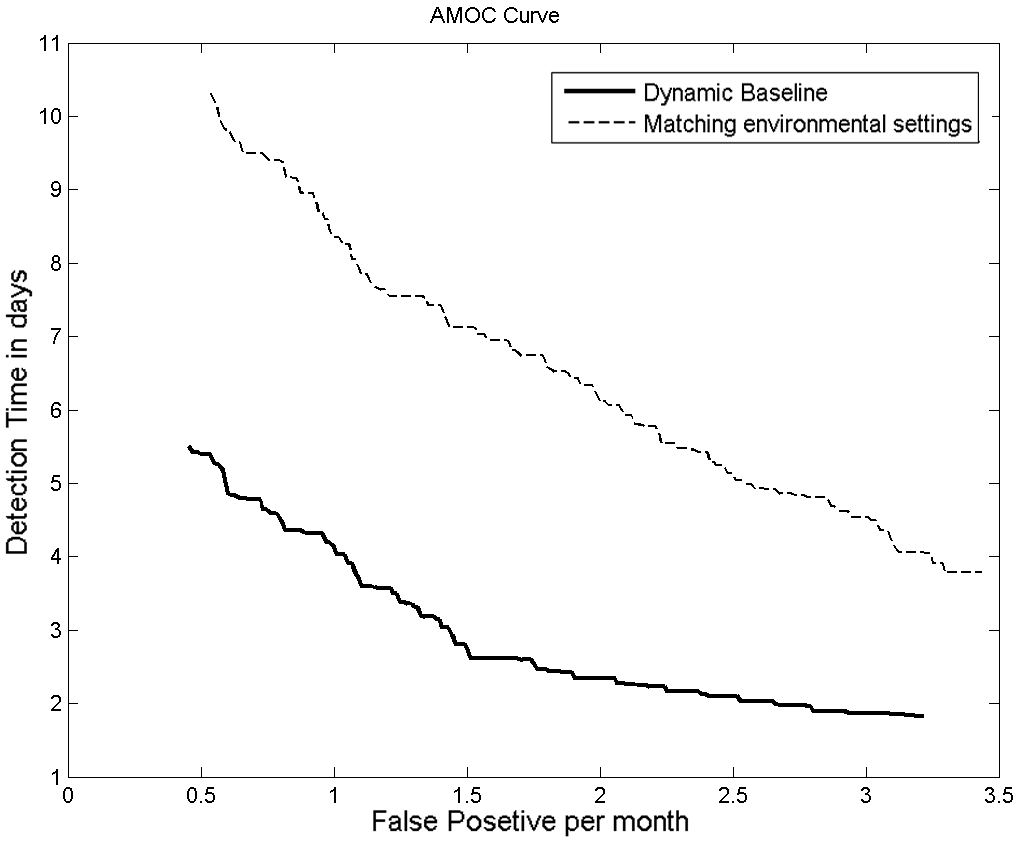}
 \end{center}
 \caption{Dynamic baseline vs. Environmental matching baseline} \label{fig:env}
\end{figure}

\begin{figure}
 \begin{center}
	\includegraphics[width=8cm]{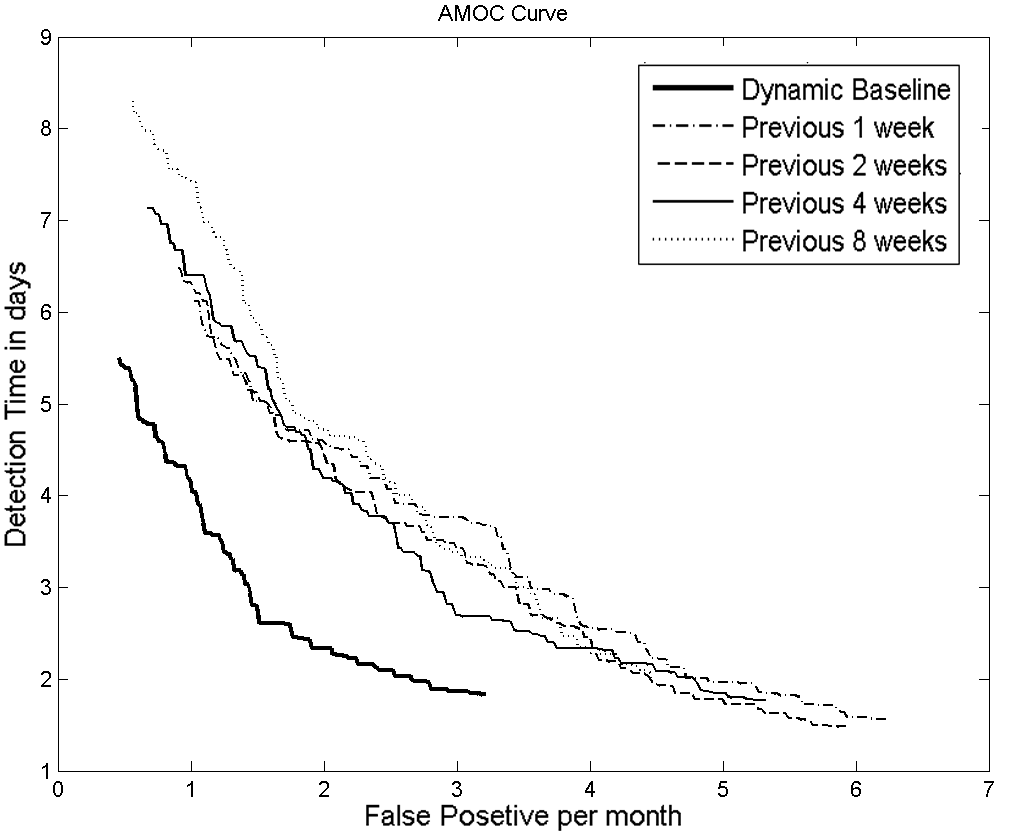}
 \end{center}
 \caption{Dynamic baseline vs. Historical baseline } \label{fig:history}
\end{figure}

We compare three scenarios for baseline creation: 1) from historical set \textit{without} respect to the environmental setting; 2) from historical set \textit{with} respect to the environmental setting; and 3) Dynamic baseline tensor (our strategy). In the first scenario we compare the recent data with historical data without considering the environmental setting. For instance, we compare the recent data with data of last one week or the last eight weeks. In the second scenario we take reference data from the matched environmental setting of the day. For instance, if the environmental setting of recent day is 4112 we search in historical set for those $Space \times Features$ matrices whose corresponding environmental setting is 4112. In the third scenario (our method), we create the baseline tensor from matched environmental setting, but we give more importance to the most dominant environmental setting and more recent data. 

Figure \ref{fig:env} and \ref{fig:history} compares the obtained performance through these different strategies. Figure \ref{fig:history} illustrates the comparison of the first scenario versus the third scenario and Figure \ref{fig:env} compares the performance of the second scenario versus the third scenario. The results reveal that our dynamic tensor creation strategy outperforms the first and second scenarios. The reason of this good performance is related to this that our approach makes a batter inference for unseen environmental setting. This approach is also robust to the noises and therefore provides a higher quality baseline reference.

\section{Conclusion and future works} \label{sec:conclusion}

We propose a novel approach based on eigenspace techniques for early detection of events from complex data streams in syndromic surveillance. The purpose of this work is to reduce the false alarm rate of the state-of-the-art early detection methods. The experimental evaluation results on benchmark data sets shows that the proposed approach provides a better overall performance versus state-of-the-art. Our approach while maintains the detection delay in a reasonable level improves the false alarm rate to a considerable extent. While top-down approaches look for changes in higher level feature space and bottom-up approaches track changes in the low-level feature space, we introduce a novel methodology based on eigenspace and tensor decomposition techniques that track changes both in high level and the dimension level. The overall changes in the system appear in the eigenvalue and a change in the dimensions appears in the eigenvectors. Such dimension-based strategy is very helpful in some applications such as disease outbreak where the spatial dimension gets very important. However, using such methods makes sense when data contains further dimensions (e.g. Space and time). In other words, the competitive part of our approach is its dimension-based change tracking which is valid only for multidimensional (multiway) data. 

A challenge to the future research is to utilize EigenEvent in a real-world problem and evaluate its performance in the practice. This was one of our main limitations in this research. Unfortunately, there is no public real-world data available with ground truth for syndromic surveillance research. Most of bio-surveillance programs also correspond to the governmental sections where gaining data in most of the time is impossible. Even if we access to real data, the period of outbreaks or events is not specified in that. There is a recent developed simulator \cite{lotze2007simulating} that simulate multivariate syndromic time series and outbreak signatures. However, since this simulator does not support the spatial dimension as the future work we are going to adapt it for this purpose and perform more experiments based on the new simulated data sets. We also intend to study the computational performance of the algorithm using incremental and streaming tensor decomposition techniques \cite{sun2008incremental} which are more appropriate for large-scale data sets. 

\subsubsection*{Complements.} The MATLAB code and data sets are available online via http://fanaee.com/research/EigenEvent.

\subsubsection*{Acknowledgments.} This research was supported by the Projects NORTE-07-0124-FEDER-000059 and NORTE-07-0124-FEDER-000056 which is financed by the North Portugal Regional Operational Program (ON.2 - O Novo Norte), under the National Strategic Reference Framework (NSRF), through the European Regional Development Fund (ERDF), and by national funds, through the Portuguese funding agency, Fundação para a Ciência ea Tecnologia(FCT). Authors also acknowledge the support of the European Commission through the project MAESTRA (Grant Number ICT-750 2013-612944).

\bibliographystyle{chicago}
\bibliography{ref}

\begin{table}
\begin{center}
\caption{Number of False Alarms per year for 100 datasets averaged for 231 p-values (0.020,0.021,..., 0.250)}\label{tab:falsereates}
		\begin{tabular}{|p{0.9cm}|p{1.2cm}|p{1.2cm}|p{1.2cm}|p{1.2cm}||p{0.9cm}|p{1.2cm}|p{1.2cm}|p{1.2cm}|p{1.2cm}|}
    \hline

   dataset & WS2.0 & WS2.5 & WS3.0 & EigenEv& dataset & WS2.0 & WS2.5 & WS3.0 & EigenEv  \\ \hline
	
1&70.013&51.8571&57.8485&\textbf{37.8398}&51&54.9827&36.1602&41.013&\textbf{24.7835}\\ \hline
2&0&0&0&0&52&89.7922&\textbf{51.29}&59.9481&51.697\\  \hline
3&\textbf{0.4416}&2.2251&1.8182&1.4545&53&89.2078&62.1991&61.2727&\textbf{46.3074}\\ \hline
4&70.4719&54.3896&53.5368&\textbf{30.71}&54&49.0952&31.7403&33.4762&\textbf{10.4805}\\ \hline
5&69.5974&40.5108&39.5758&\textbf{29.0823}&55&47.4156&42.1991&48.3896&\textbf{29.2381}\\ \hline
6&93.7792&59.0779&62.0346&\textbf{37.7056}& 56&108.3766&73.5108&71.3333&\textbf{40.5498}\\ \hline
7&72.5628&50.5584&62.1082&\textbf{35.3203}& 57&59.6017&36.7489&38.3117&\textbf{20.0866}\\ \hline
8&89.5628&71.039&79.2771&\textbf{50.0303}& 58&25.1299&20.4589&22.1948&\textbf{13.7706}\\ \hline
9&26.3506&17.684&18.4632&\textbf{8.2468}& 59&90.4372&41.8658&41.6797&\textbf{40.0779}\\ \hline
10&70.0563&53.8831&49.5584&\textbf{34.7532}& 60&79.0996&51.2208&52.3074&\textbf{50.7013}\\ \hline
11&89.6104&63.2251&53.2771&\textbf{48.3636}& 61&70.4329&44.29&50.4286&\textbf{29.316}\\ \hline
12&89.3896&71.3333&75.1342&\textbf{48.4805}& 62&31.5455&19.1645&19.7879&\textbf{9.342}\\ \hline
13&82.8442&51.0996&49.0346&\textbf{40.0346}& 63&59.3463&42.632&45.2511&\textbf{33.2597}\\ \hline
14&35.7013&25.2597&25.4762&\textbf{17.632}& 64&72.6407&48.4892&51.9221&\textbf{37.29}\\ \hline
15&80.7619&44.6926&43.6407&\textbf{34.7879}& 65&29.2251&19.684&20.2857&\textbf{9.2684}\\ \hline
16&58.8745&42.7056&42.5195&\textbf{25.7273}& 66&3.2771&2.3723&3.987&\textbf{0.8312}\\ \hline
17&5.5714&4.2208&6.6753&\textbf{1.3939}& 67&39.8398&22.6537&22.9351&\textbf{10.2468}\\ \hline
18&62.0433&42.0606&44.0346&\textbf{16.2338}& 68&30.5844&13.4286&15.8052&\textbf{7.4286}\\ \hline
19&55.1472&38.7532&45.6017&\textbf{32.9567}& 69&9.0476&8.0087&10.5931&\textbf{2.9004}\\ \hline
20&31.8225&31.3896&30.4675&\textbf{17.3463}& 70&1.0952&\textbf{0}&\textbf{0}&\textbf{0}\\ \hline
21&17.3074&9.9221&11.8831&\textbf{8.3117}& 71&7.961&7.7489&9.5281&\textbf{2.7186}\\ \hline
22&42.4502&22.3506&20.9091&\textbf{11.1775}& 72&9.974&9.645&7.7835&\textbf{2.9697}\\ \hline
23&10.9913&7.2251&8.6667&\textbf{0.9524}& 73&35.7749&28.3939&27.0476&\textbf{16.6667}\\ \hline
24&29.4199&20.2338&26.2035&\textbf{10.1255}& 74&18.5801&13.7706&13.0563&\textbf{6.3853}\\ \hline
25&35.1905&25.3377&22.7706&\textbf{14.5455}& 75&78.0433&44.7316&56.9134&\textbf{35.2424}\\ \hline
26&84.2597&54.7229&57.4589&\textbf{50.0909}& 76&73.1082&48.8095&\textbf{48.6753}&52.2165\\ \hline
27&9.987&9.8052&11.2641&\textbf{4.4372}& 77&8.8615&8.2944&8.3983&\textbf{0.2381}\\ \hline
28&20.3896&16.4632&14.7316&\textbf{3.1732}& 78&29.5238&21.5108&12.5455&\textbf{10.697}\\ \hline
29&0&0&0&0& 79&76.4675&53.1732&55.961&\textbf{50.9437}\\ \hline
30&93.2468&65.0736&69.4156&\textbf{48.4199}& 80&48.5152&23.4459&23.5325&\textbf{15.3074}\\ \hline
31&47.8701&35.5108&36.8398&\textbf{32.8528}& 81&\textbf{4.2338}&10.303&9.1645&4.5801\\ \hline
32&74.7013&60.7835&70.7576&\textbf{40.3939}& 82&38.316&22.0779&21.9481&\textbf{10.3377}\\ \hline
33&\textbf{0}&\textbf{0}&0.6537&\textbf{0}& 83&85.9004&57.8052&57.1775&\textbf{40.6797}\\ \hline
34&71.7229&40.6623&40.9437&\textbf{20.0866}& 84&8.3074&4.7965&4.5152&\textbf{0.7749}\\ \hline
35&88.7706&54.7489&\textbf{48.5152}&52.1472& 85&20.303&14.8095&12.4978&\textbf{3.1255}\\ \hline
36&61.2468&50.4719&40.6407&\textbf{39.8268}& 86&77.6061&42.0996&49.8745&\textbf{35.7532}\\ \hline
37&31.0866&15.4719&\textbf{12.961}&16.8918& 87&22.5671&18.0173&22.4459&\textbf{8.658}\\ \hline
38&31.6104&23.2771&27.3247&\textbf{18.7662}& 88&8.3723&6.5238&7.0303&\textbf{1.316}\\ \hline
39&95.3203&65.1385&73.9913&\textbf{46.8831}& 89&57.5108&40.0606&51.1212&\textbf{21.7229}\\ \hline
40&33.3939&18.9307&23.9177&\textbf{15.8874}& 90&67.71&46.0433&58.1385&\textbf{24.2381}\\ \hline
41&65.0909&39.8918&39.2208&\textbf{31.29}& 91&67.7749&48.6797&43.1169&\textbf{20.7056} \\ \hline
42&77.7359&43.7273&49.8528&\textbf{24}& 92&68.316&55.5541&56.2035&\textbf{34.9221}\\ \hline
43&30.9264&20.4719&20.5887&\textbf{10.4762}&93&66.987&48.7013&47.1039&\textbf{28.5022}\\ \hline
44&63.0303&46.039&48.8918&\textbf{36.4416}& 94&28.697&24.1688&27.7056&\textbf{9.7576}\\ \hline
45&81.7879&47.7792&49.3506&\textbf{45.2771}& 95&69.4632&48.1169&47.2944&\textbf{33.2814}\\ \hline
46&84.7749&65.7662&67.8355&\textbf{41.4589}& 96&34.0216&16.8528&18.2121&\textbf{12.4502}\\ \hline
47&8.3377&11.1688&9.684&\textbf{0.5108}& 97&4.0996&6.1818&7.5498&\textbf{1.619}\\ \hline
48&43.6234&19.7879&29.7489&\textbf{15.4242}& 98&34.0476&18.4372&15.0087&\textbf{14.7403}\\ \hline
49&26.5195&\textbf{14.1861}&18.1169&15.4286& 99&56.0909&30.987&42.8485&\textbf{22.7186}\\ \hline
50&33.4329&31.6797&35.3117&\textbf{16.5671}& 100&66.7922&46.4242&50.5844&\textbf{28.9394}\\ \hline

  \end{tabular}
\end{center}
\end{table}

\begin{table}
\begin{center}
\caption{Detection Delay (in days) for 100 datasets averaged for 231 p-values (0.020,0.021,..., 0.250)}\label{tab:delays}
		\begin{tabular}{|p{0.9cm}|p{1.2cm}|p{1.2cm}|p{1.2cm}|p{1.2cm}||p{0.9cm}|p{1.2cm}|p{1.2cm}|p{1.2cm}|p{1.2cm}|}
    \hline

   dataset & WS2.0 & WS2.5 & WS3.0 & EigenEv& dataset & WS2.0 & WS2.5 & WS3.0 & EigenEv  \\ \hline
	
1&\textbf{1.3463}&2&\textbf{1.3463}&2&51&\textbf{1}&\textbf{1}&\textbf{1}&3.5325\\ \hline
2&\textbf{1.0433}&1&1&1.2987&52&\textbf{1}&\textbf{1}&\textbf{1}&2.0693\\ \hline
3&\textbf{2}&\textbf{2}&\textbf{2}&12.9394&53&\textbf{3.7835}&8.7532&7.7706&4.8312\\ \hline
4&1&1&1&1&54&1&1&1&1\\ \hline
5&\textbf{1}&\textbf{1}&\textbf{1}&2.7273&55&1&1&1&1\\ \hline
6&\textbf{4.7056}&6.2035&5.2511&6.0346&56&1&1&1&1\\ \hline
7&\textbf{1}&\textbf{1}&\textbf{1}&1.5065&57&1&1&1&1\\ \hline
8&1.2251&\textbf{1}&\textbf{1}&1.6104&58&1&1&1&1\\ \hline
9&1.5628&1.2771&1.1688&\textbf{1}&59&5.0909&4.4892&6.1212&11.0606\\ \hline
10&\textbf{1}&9.1905&\textbf{1}&\textbf{1}&60&3.039&2&\textbf{1.5628}&5.4286\\ \hline
11&\textbf{1}&\textbf{1}&\textbf{1}&1.5974&61&\textbf{1.9957}&2&2&2\\ \hline
12&1&1&1&1&62&1&1&1&1\\ \hline
13&4.2078&\textbf{1}&\textbf{1}&7.8139&63&\textbf{1}&2&\textbf{1}&\textbf{1}\\ \hline
14&1&1&1&1&64&4.5065&13.9913&\textbf{1.7316}&2.026\\ \hline
15&2.3593&\textbf{1}&\textbf{1}&1.7316&65&\textbf{4.0779}&7.6364&10.4156&8.4502\\ \hline
16&1.039&\textbf{1}&\textbf{1}&1.0996&66&2&\textbf{1.3463}&2&1.5758\\ \hline
17&\textbf{1}&\textbf{1}&\textbf{1}&1.5065&67&2.8095&\textbf{1}&\textbf{1}&1.3506\\ \hline
18&\textbf{1}&\textbf{1}&\textbf{1}&1.026&68&1&1&1&1\\ \hline
19&1.3074&\textbf{1}&1.2338&2&69&\textbf{1}&\textbf{1}&\textbf{1}&3.8139\\ \hline
20&1&1&1&1&70&\textbf{1}&\textbf{1}&\textbf{1}&2.5931\\ \hline
21&1&1&1&1&71&1&1&1&1\\ \hline
22&1&1&1&1&72&6.5368&\textbf{3}&\textbf{3}&11.5455\\ \hline
23&1&1&1&1&73&\textbf{4}&7.5758&6.3463&5.3333\\ \hline
24&1&1&1&1&74&1.2597&\textbf{1}&\textbf{1}&\textbf{1}\\ \hline
25&1&1&1&1&75&\textbf{1}&2.1255&\textbf{1}&\textbf{1}\\ \hline
26&2.5541&2&\textbf{1.2035}&5.0649&76&\textbf{1.2121}&2&1&1.2251\\ \hline
27&1.4329&\textbf{1}&\textbf{1}&\textbf{1}&77&1.9091&\textbf{1}&\textbf{1}&2.2468\\ \hline
28&1&1&1&1&78&10.0779&\textbf{7.4632}&13.7403&11.2294\\ \hline
29&1&1&1&1&79&\textbf{4.2554}&8.1039&4.961&8.7532\\ \hline
30&8.6883&2.0823&\textbf{2}&4.2338&80&1&1&1&1\\ \hline
31&1&1&1&1&81&1&1&1&1\\ \hline
32&5.4242&1.2597&\textbf{1}&2.3506&82&1.4502&\textbf{1}&\textbf{1}&\textbf{1}\\ \hline
33&1&1&1&1&83&6.6753&\textbf{4.5022}&7.355&5.6147\\ \hline
34&1&1&1&1&84&1.3463&\textbf{1}&\textbf{1}&\textbf{1}\\ \hline
35&3.6883&\textbf{1.329}&1.8788&5.6623&85&1&1&1&1\\ \hline
36&\textbf{2.2597}&6.4935&7.1126&10.1082&86&\textbf{1.7835}&2&2&1.974\\ \hline
37&1&1&1&1&87&2.9134&2.645&1&4.0823\\ \hline
38&1&1&1&1&88&\textbf{1}&\textbf{1}&\textbf{1}&3.0823\\ \hline
39&5.7186&6.4242&10.0649&10.7273&89&1&1&1&1\\ \hline
40&1.0476&\textbf{1}&\textbf{1}&\textbf{1}&90&1&1&1&1\\ \hline
41&1&1&1&1&91&2.1732&1.0866&\textbf{1}&7.1602\\ \hline
42&1.5628&\textbf{1}&\textbf{1}&\textbf{1}&92&6.8095&\textbf{3.4762}&5.1429&11.1602\\ \hline
43&\textbf{1}&10.0779&\textbf{1}&\textbf{1}&93&1.5628&1.3463&\textbf{1}&2.329\\ \hline
44&2.329&\textbf{1}&\textbf{1}&1.961&94&1.2468&1.329&\textbf{1.1299}&1.9827\\ \hline
45&6.3074&2.3853&2.039&\textbf{1}&95&1.4502&\textbf{1}&\textbf{1}&\textbf{1}\\ \hline
46&4.5281&\textbf{1}&\textbf{1}&6.8831&96&1&1&1&1\\ \hline
47&\textbf{7.3506}&10.2944&7.9913&12.9957&97&1&1&1&1\\ \hline
48&1.2727&\textbf{1}&\textbf{1}&1.3593&98&3.645&\textbf{1}&\textbf{1}&1.3377\\ \hline
49&4.8268&3.3463&2.3463&\textbf{1}&99&1&1&1&1\\ \hline
50&1&1&1&1&100&\textbf{1}&\textbf{1}&\textbf{1}&1.9567\\ \hline

  \end{tabular}
\end{center}
\end{table}

\end{document}